\documentclass[conference]{IEEEtran}

\IEEEoverridecommandlockouts                              

\usepackage[utf8]{inputenc}
\usepackage{graphics} 
\usepackage{epsfig} 
\usepackage{amsmath} 
\usepackage{amsfonts}
\usepackage[breaklinks=true]{hyperref}
\usepackage{booktabs}
\usepackage[font=small]{caption}
\usepackage{balance}
\usepackage{amsfonts}
\usepackage{tabularx}
\usepackage{tikz}
\usepackage{nicematrix}
\usepackage{graphicx}
\usepackage{comment}
\usepackage{cite}
\let\labelindent\relax
\usepackage[shortlabels]{enumitem}
\usepackage{adjustbox}
\usepackage{dblfloatfix}

\newcommand{\PreserveBackslash}[1]{\let\temp=\\#1\let\\=\temp}
\newcolumntype{C}[1]{>{\PreserveBackslash\centering}p{#1}}
\newcommand\Tstrut{\rule{0pt}{2.8ex}}         
\newcommand\Bstrut{\rule[-2ex]{0pt}{0pt}}   
\newcommand\Tstruttwo{\rule{0pt}{3.2ex}}         
\newcommand\Bstruttwo{\rule[-2.4ex]{0pt}{0pt}}   
\newcommand\picdims[4][]{%
  \setbox0=\hbox{\includegraphics[#1]{#4}}%
  \clipbox{.5\dimexpr\wd0-#2\relax{} %
           .5\dimexpr\ht0-#3\relax{} %
           .5\dimexpr\wd0-#2\relax{} %
           .5\dimexpr\ht0-#3\relax}{\includegraphics[#1]{#4}}}

\usepackage{titlesec}
\setlist[itemize]{noitemsep, ,nolistsep,topsep=0pt}
\setlist[enumerate]{noitemsep,nolistsep, topsep=0pt}
\title{\LARGE \bf
Supermarket-6DoF: A Real-World  Grasping Dataset and Grasp Pose Representation Analysis
}

\author{Jason Toskov$^{1}$ and Akansel Cosgun$^{2}$%
\thanks{$^{1}$ Swiss Federal Institute of Technology Lausanne (EPFL), Switzerland}
\thanks{$^{2}$ Deakin University, Australia}
}

\begin{document}

\maketitle
\thispagestyle{empty}
\pagestyle{empty}

\begin{abstract}
We present Supermarket-6DoF, a real-world dataset of 1500 grasp attempts across 20 supermarket objects with publicly available 3D models. Unlike most existing grasping datasets that rely on analytical metrics or simulation for grasp labeling, our dataset provides ground-truth outcomes from physical robot executions. Among the few real-world grasping datasets, wile more modest in size, Supermarket-6DoF uniquely features full 6-DoF grasp poses annotated with both initial grasp success and post-grasp stability under external perturbation. We demonstrate the dataset's utility by analyzing three grasp pose representations for grasp success prediction from point clouds. Our results show that representing the gripper geometry explicitly as a point cloud achieves higher prediction accuracy compared to conventional quaternion-based grasp pose encoding.


\end{abstract}


\section{Introduction}

Robotic grasping research aims to enable robots to manipulate a wide variety of real-world objects effectively. However, achieving standardized experimental practices remains a significant challenge, as researchers historically relied on objects readily available in their laboratories, limiting meaningful benchmarking efforts. To address this, the robotics community has developed standardized grasping datasets for more consistent evaluations and learning-based approaches, as shown in Table \ref{tab:related_grasps}

There are three fundamental ways to label a grasp attempt: human annotation, synthetically estimated via analytical metrics and trial and error execution. Early datasets like Cornell \cite{5152709} as well as Zhang et al. \cite{zhang2019roi} relied on manual labeling of bounding boxes in real images to represent top-down grasp attempts with parallel grippers. While Murali et al. \cite{murali2020taskgrasp} extended this approach to crowdsourced labeling of simulated 6-DoF grasps, manual annotation has seen limited adoption due to its labor-intensive nature.

A more scalable approach involves estimating grasp success through analytical grasp quality metrics without physical execution of the grasp. The Columbia grasp database \cite{5152709} and Drögemüller et al. \cite{9659350} evaluate top-down grasps in simulation using epsilon grasp quality and force closure metrics, respectively. Dex-Net 2.0 \cite{DBLP:conf/rss/MahlerLNLDLOG17} and subsequent works \cite{9681218,morrison2020egad} expanded this methodology to encompass diverse grasp configurations beyond top-down grasps. However, these analytical metrics often fail to capture the complexities of real-world interactions, as demonstrated by Kappler et al. \cite{7139793}. This limitation led to the adoption of simulation-based grasp execution for more direct quality assessment, both for top-down \cite{Depierre2018JacquardAL} and 6 DoF grasps \cite{Veres2017AnIS,EppnerISRR2019,mousavian2019graspnet,acronym2020,zhai2022monograspnet}. Despite these advances, the sim-to-real gap persists, as simulated environments cannot fully replicate the physical properties and dynamics of the real-world. Some researchers have attempted to bridge this gap by combining real sensor data with analytical metrics \cite{9156992} or physical simulation \cite{zhai2022monograspnet}.

\begin{figure}[t!]
    \centering
    \includegraphics[trim={450 35 50 230}, clip, width=0.238\textwidth]{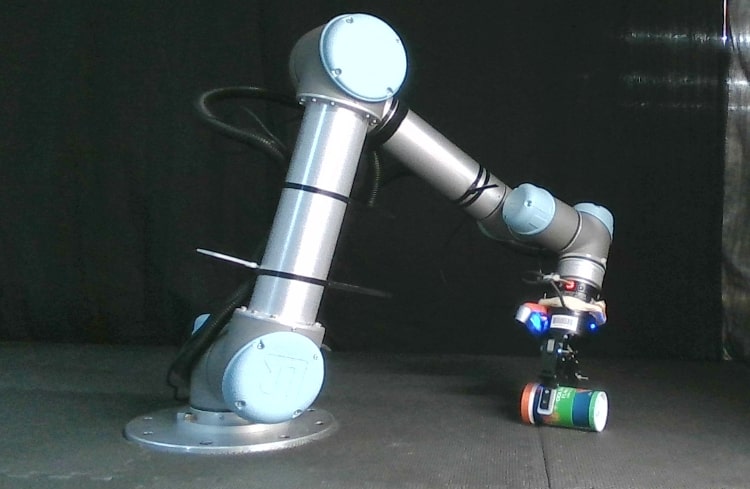}
    \includegraphics[trim={450 35 50 230}, clip, width=0.238\textwidth]{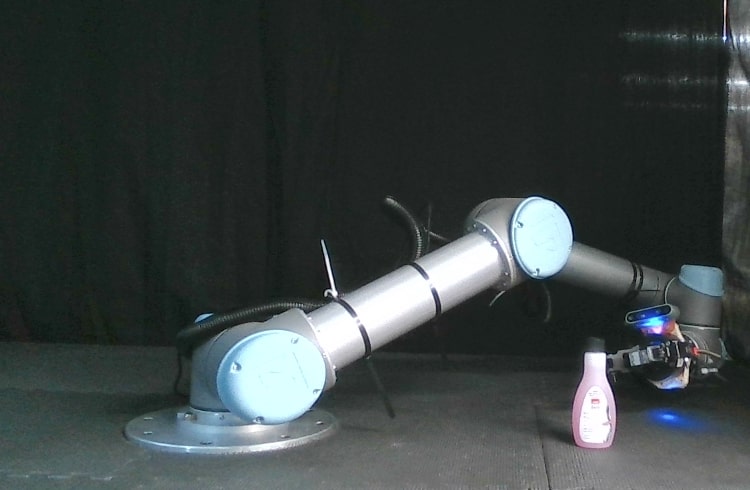}
    \caption{We present a real-world grasping dataset with 1500 grasp attempts on 20 supermarket objects, featuring 6-DoF grasp attempts.}
    \label{fig:intro}
\end{figure}

To overcome simulation limitations, some researchers have developed grasping datasets with physical grasp execution on real robots, such as Pinto et al. \cite{Pinto2015SupersizingSL} and Levine et al. \cite{doi:10.1177/0278364917710318}. These datasets, however, are contrained to 4 degrees of freedom (4DoF) grasps, specifying only the end-effector's position (x, y, z) and a single rotation (typically around the vertical z-axis). This restriction suffices for simple pick-and-place tasks but fails to address complex manipulation scenarios requiring arbitrary object orientations. While simulation-based datasets like ACRONYM \cite{acronym2020} and MonoGraspNet \cite{zhai2022monograspnet} have incorporated full 6-DoF grasps, no real-world dataset currently features comprehensive 6-DoF grasp attempts, representing a critical gap in existing grasping datasets.

Furthermore, traditional datasets often employ binary success labels (successful/failed) based on initial execution. However, robust evaluation requires testing grasp stability under perturbations. While robustness testing has been integrated into simulation-based datasets \cite{7139793,EppnerISRR2019,acronym2020}, real-world grasping datasets have yet to incorporate such stability assessments.

To address these limitations, we present the Supermarket-6DoF dataset\footnote{\href{https://lachlanchumbley.github.io/Monash-ColesGraspingDataset/}{https://lachlanchumbley.github.io/Monash-ColesGraspingDataset/}.}, comprising 1,500 real-world grasp attempts performed on 20 supermarket objects. Our dataset features full 6-DoF end-effector poses, enabling diverse and adaptable grasping strategies. Additionally, we investigate the impact of different gripper pose representations on grasp success prediction, providing insights into effective feature engineering for learning-based grasp synthesis \cite{newbury2023deep}.

\begin{table}
\setlength{\tabcolsep}{3pt}
    \centering
    \begin{tabular}{|l|c|C{0.044\textwidth}|C{0.038\textwidth}|C{0.07\textwidth}|C{0.047\textwidth}|} 
          \hline \textbf{Grasping Dataset}  & \textbf{6DoF} & \textbf{Num grasps} & \textbf{Sim/ Real} & \textbf{Labels} & \textbf{Robust check} \\ \hline
          Cornell \cite{5980145}  & & 8k & Real & Human &  \\ \hline
          Zhang et al. \cite{zhang2019roi}  & & 100k & Real & Human &  \\ \hline
          Murali et al. \cite{murali2020taskgrasp}  &\checkmark & 250k & Sim & Human &  \\ \hline
          Columbia \cite{5152709}  & &  238k & Sim & Analytic &  \\ \hline
          Drögemüller \cite{9659350}    && 15k & Sim & Analytic &  \\ \hline
          Grasp-Anything \cite{vuong2023grasp} & & 600M & Sim & Analytic & \\ \hline
          Grasp-Anything-6D\cite{nguyen2025language} & \checkmark & 200M & Sim & Analytic & \\ \hline
          REGRAD \cite{9681218}  &\checkmark & 118k & Sim & Analytic &  \\ \hline
          Dex-Net 2.0 \cite{DBLP:conf/rss/MahlerLNLDLOG17}   & \checkmark & 6.7M & Sim & Analytic &  \\ \hline
          EGAD! \cite{morrison2020egad} & \checkmark & 1M & Sim & Analytic & \\ \hline
          GraspNet-1Billion \cite{9156992}  &\checkmark & 1.2B & Both & Analytic &  \\ \hline   
          Kappler et al\cite{7139793}   & \checkmark  & 300k & Sim & Analytic, Trials & \checkmark \\ \hline
          Sim-Suction \cite{li2023sim} & \checkmark & 3M & Sim & Analytic, Trials & \\ \hline
          Jacquard \cite{Depierre2018JacquardAL}   & & 1.1M & Sim & Trials &  \\ \hline
          Veres et al. \cite{Veres2017AnIS}  &\checkmark & 50k & Sim & Trials &  \\ \hline
          Billion Ways \cite{EppnerISRR2019}  &\checkmark & 1B & Sim & Trials & \checkmark \\ \hline          
          6DoF GraspNet\cite{mousavian2019graspnet} & \checkmark & 7M & Sim & Trials &  \\ \hline 
          ACRONYM \cite{acronym2020}  &\checkmark & 17M & Sim & Trials & \checkmark \\ \hline
          MonoGraspNet \cite{zhai2022monograspnet}   &\checkmark & 20M & Both & Trials &  \\ \hline      
          Pinto et al. \cite{Pinto2015SupersizingSL}   & & 50k & Real & Trials &  \\ \hline
          Levine et al. \cite{doi:10.1177/0278364917710318}  & & 800k & Real & Trials &  \\ \hline
          \textbf{Supermarket-6DoF}  &\checkmark & \textbf{1.5k} & \textbf{Real} & \textbf{Trials} & \checkmark \\ \hline
    \end{tabular}
\caption{Grasping datasets in the literature}
\label{tab:related_grasps}
\end{table}

\section{Supermarket-6DoF Grasping Dataset}
\label{sec:grasping_ds}

\subsection{Data Overview}

The Supermarket-6DoF dataset provides comprehensive data for each grasp attempt, including:

\begin{itemize}
\item \textbf{Sensor Data}: A single-view eye-in-hand RGB image, a depth image, and the corresponding point cloud of the grasp scene.
\item \textbf{Grasp Pose}: The 6-DoF pose of the gripper (position and orientation)
\item \textbf{Grasp Annotations}: binary labels indicating:
\begin{enumerate}
    \item Grasp success: Whether the object was successfully lifted by the gripper.
    \item Grasp robustness: Whether the object remained stable under physical perturbation post-grasp.
\end{enumerate}

\end{itemize}

\subsection{Object Selection}

We selected 20 objects from the Supermarket Object Set\footnote{\href{https://lachlanchumbley.github.io/ColesObjectSet/}{https://lachlanchumbley.github.io/ColesObjectSet/}.} with the following criteria:

\begin{itemize} 
    \item \textbf{Diversity}: Objects were chosen to represent a wide range of shapes, sizes, weights, and materials. Diverse shapes, sizes, weights, and materials
    \item \textbf{Compatibility}: All objects are graspable by standard parallel-jaw grippers.
    \item \textbf{Practical Relevance}: The objects reflect real-world scenarios encountered in everyday manipulation tasks.
\end{itemize}

The object selection enables thorough evaluation of grasping algorithms, particularly for learning-based methods. Object details are provided in Table \ref{tab:per_object_results}.

\subsection{Hardware Setup}

We used a UR5 robotic arm with a Robotiq 2f-85 parallel-jaw gripper. An Intel RealSense D435 RGB-D camera with an eye-in-hand configuration was attached to the robot's wrist. While a Robotiq FT-300 force/torque sensor was attached to the setup, its readings were not utilized in the dataset. The extrinsic calibration of the camera was achieved by the easy\_handeye\footnote{\href{https://github.com/IFL-CAMP/easy_handeye}{https://github.com/IFL-CAMP/easy\_handeye}} library.

\subsection{Data Collection Methodology}
\label{subsec:grasp_data_collection}

For each object, we collected 75 grasp attempts:

\begin{itemize}
\item 25 attempts with upright object placement
\item 50 attempts with random object orientations
\end{itemize}

The data collection procedure for each grasp attempt consisted of the following steps:

\begin{enumerate}
    \item \textbf{Object Positioning}: The object was either manually placed in an upright position or dropped onto the table to create a random orientation. All object positions were constrained to a predefined workspace to ensure they remained fully scannable.
    \item \textbf{Scene capture}: A single-view scan was captured from one of four predefined scan locations. These scan locations were cycled sequentially across attempts to ensure varied perspectives. Each scan provided an RGB image, a depth image and a point cloud of the scene, transformed to the world frame.
    \item \textbf{Grasp Selection}: Using the method described in Gualtieri et al. \cite{7759114}, 500 candidate 6-DoF grasp poses, along with their estimated probability of success, were generated for the scanned point cloud. The selection strategy was based on the current grasp success ratio for that object. If the success ratio was below 80\%, a grasp pose was randomly selected from the top 250 proposals. If the success ratio was above 80\%,  a grasp pose was randomly selected from the bottom 250 proposals, with a small noise added to the pose for variability. The 80\% threshold was empirically determined to maintain a balanced dataset of successful and failed grasp attempts.
    \item \textbf{Grasp Execution}: The robot first moved the gripper to a pre-grasp pose, offset 10 cm from the selected grasp pose, before executing the grasp. The gripper was then closed around the object. If the gripper successfully closed on the object the grasp was classified as a ``success". If the gripper successfully closed on the object and lifted it, the grasp was labeled as a "success." For successful grasps, a stability test was conducted by lifting the object to a predefined height and rotating the wrist joint by $\pm90^\circ$, simulating object perturbation. If the object remained secure throughout this process, the grasp was further labeled as ``stable."
\end{enumerate}

\begin{figure*}[!t]
    \centering
    \begin{NiceTabularX}{\textwidth}{X[1,c]X[1,c]X[1,c]X[1,c]X[1,c]X[1,c]}
        \includegraphics[width=\textwidth]{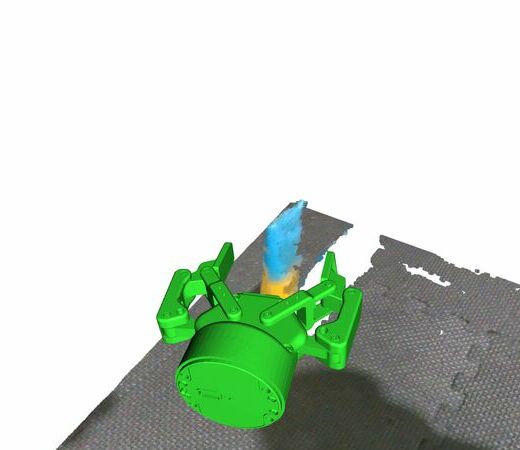}&
        \includegraphics[width=\textwidth]{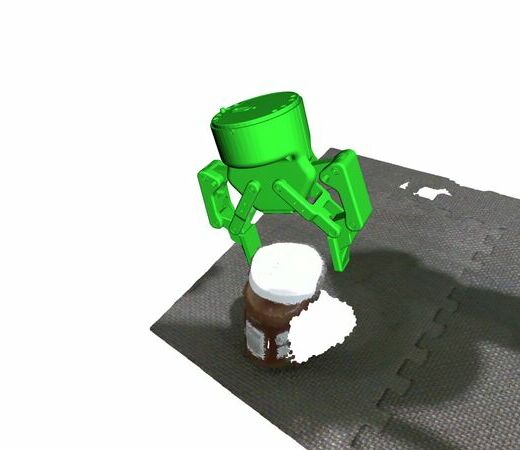}&
        \includegraphics[width=\textwidth]{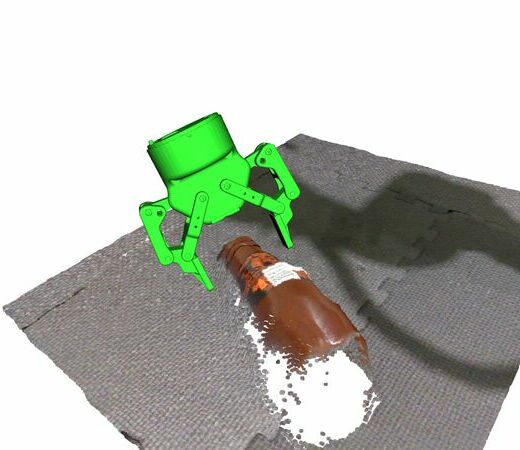}&
        \includegraphics[width=\textwidth]{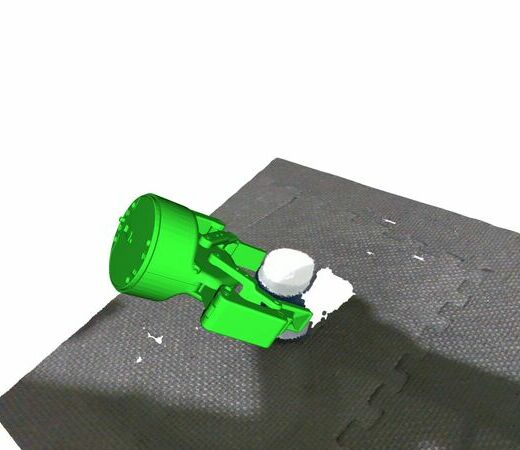}&
        \includegraphics[width=\textwidth]{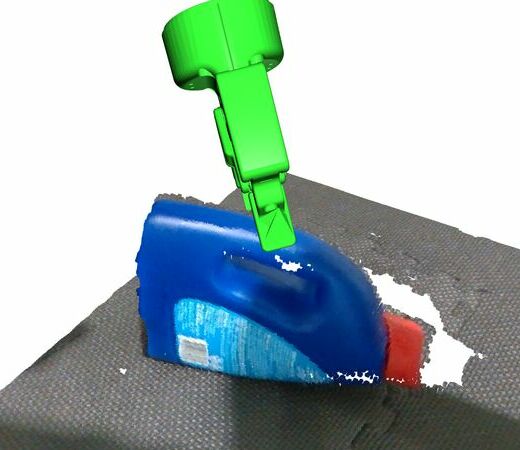}&
        \includegraphics[width=\textwidth]{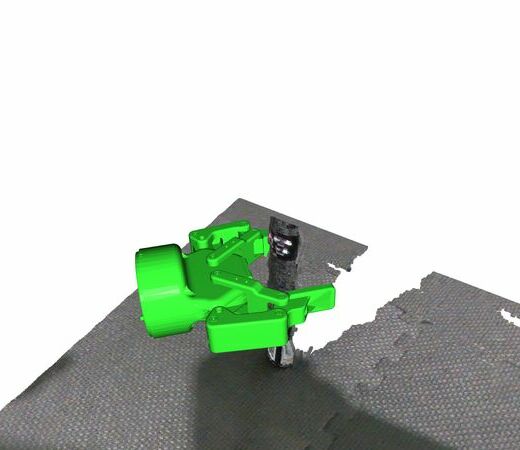}\\
        \includegraphics[width=\textwidth]{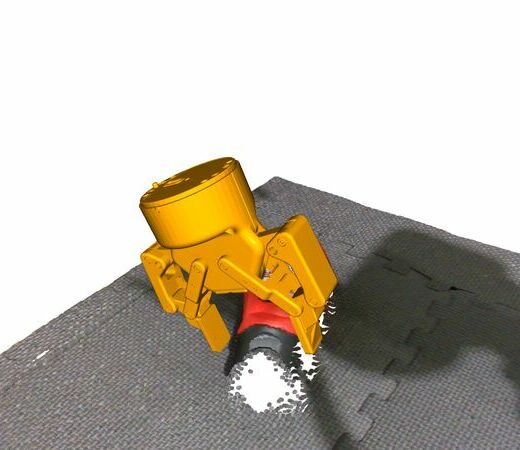}&
        \includegraphics[width=\textwidth]{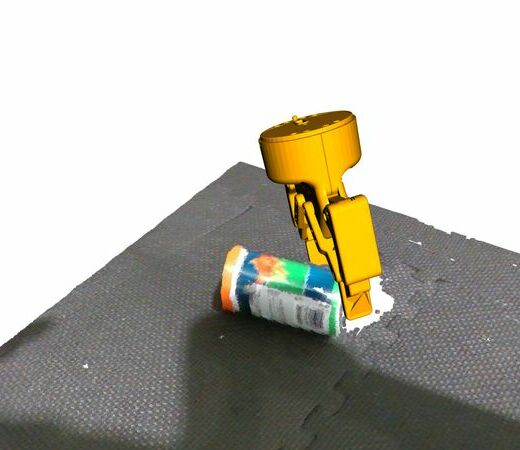}&
        \includegraphics[width=\textwidth]{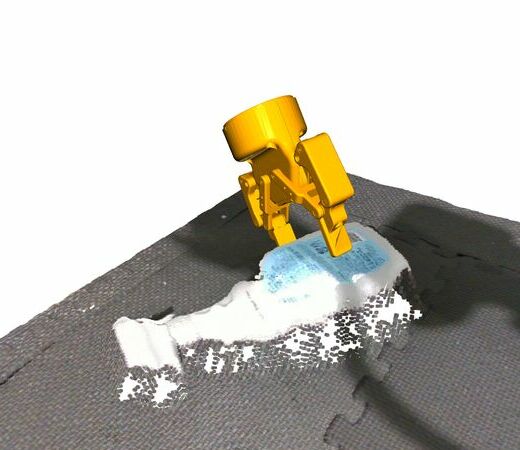}&
        \includegraphics[width=\textwidth]{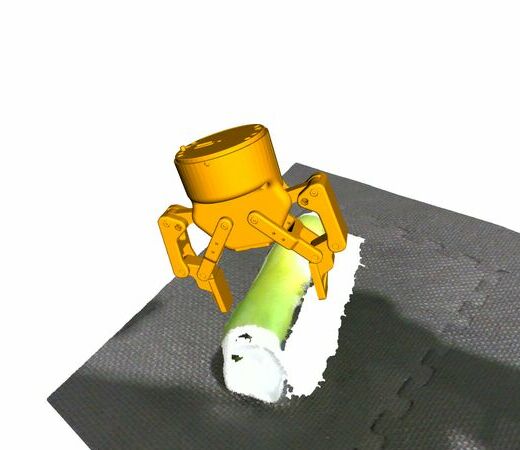}&
        \includegraphics[width=\textwidth]{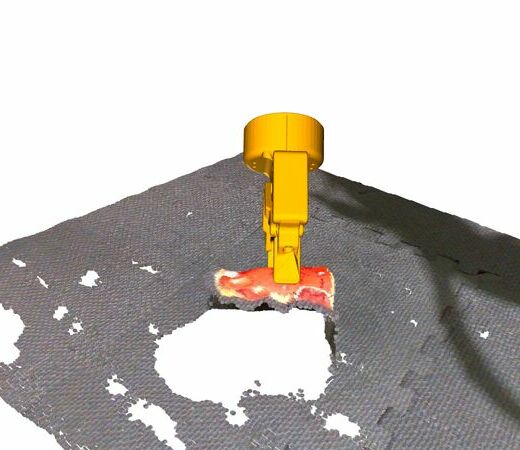}&
        \includegraphics[width=\textwidth]{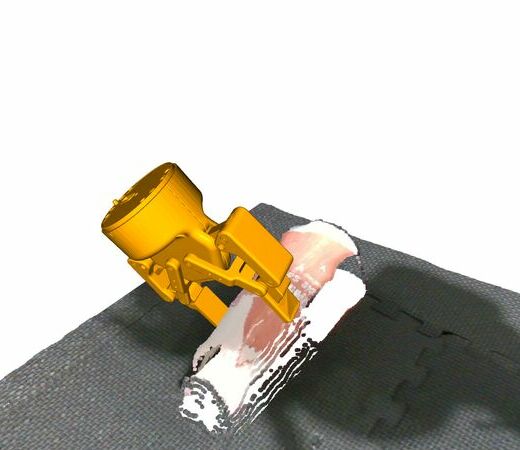}\\
        \includegraphics[width=\textwidth]{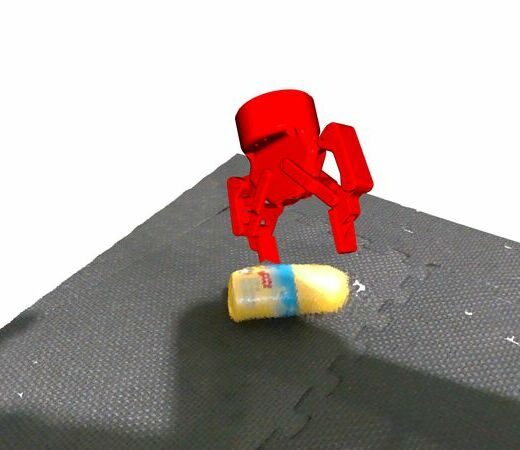}&
        \includegraphics[width=\textwidth]{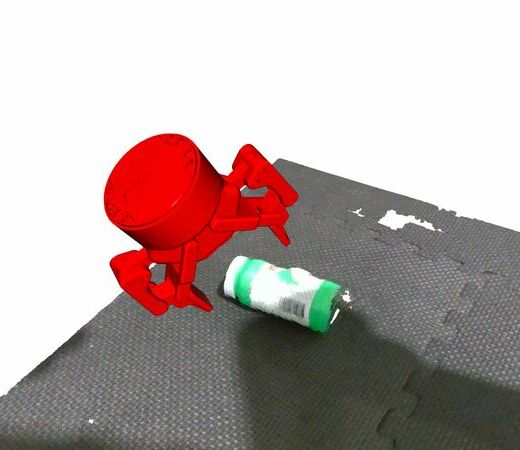}&
        \includegraphics[width=\textwidth]{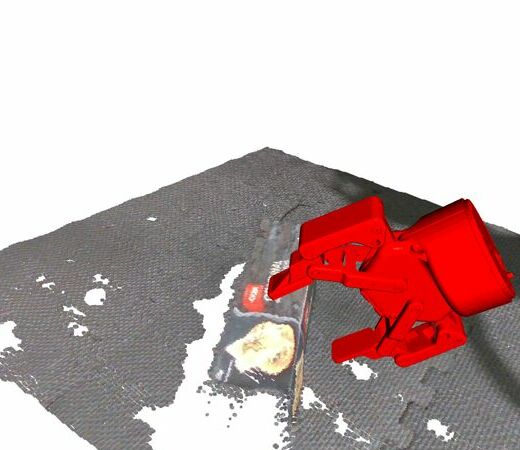}&
        \includegraphics[width=\textwidth]{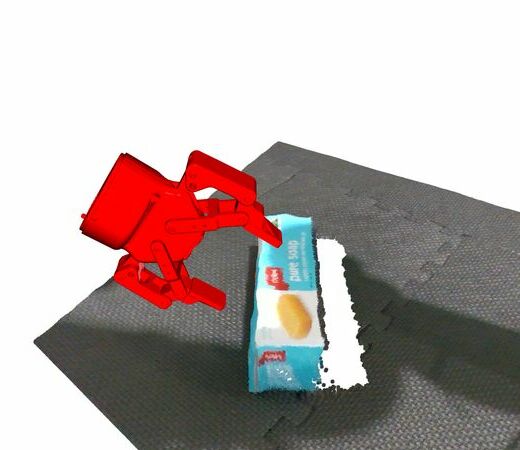}&
        \includegraphics[width=\textwidth]{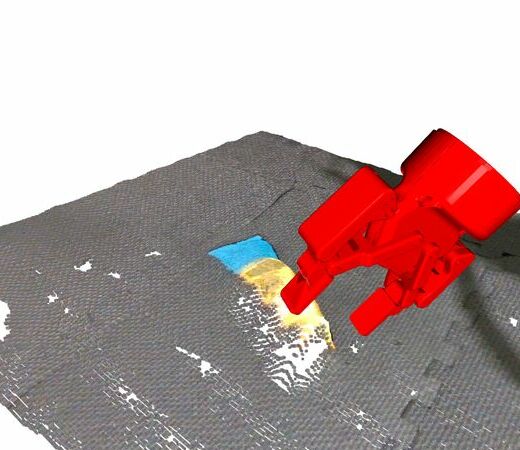}&
        \includegraphics[width=\textwidth]{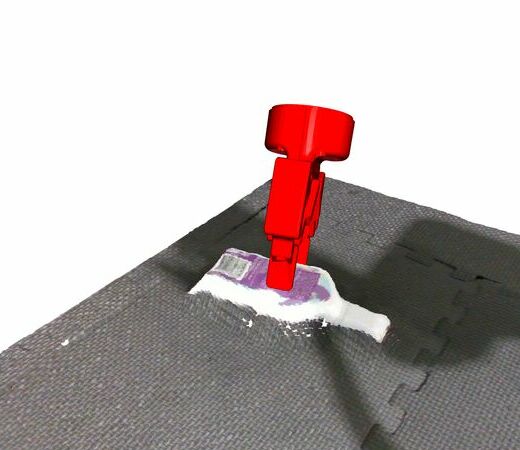}\\
    \end{NiceTabularX}
    \caption{Example grasps from the Supermarket-6DoF dataset. Green grasps (top) are stable successes, orange grasps (middle) are successes that aren't stable, and red grasps (bottom) are failures.}
    \label{fig:grasp_examples}
    \vspace{-0.3cm}
\end{figure*}

This procedure was applied to 20 supermarket objects, resulting in a total of 1,500 labeled grasp attempts. Example grasp attempts, along with their success and stability labels, are illustrated in Figure \ref{fig:grasp_examples}.

\section{Grasp Success Prediction}
\label{sec:success_prediction}

Grasp success prediction is the task of determining whether a given grasp pose will result in a successful grasp. This task is a critical component of grasping algorithms and has been addressed using various techniques in the literature. Past approaches have employed support vector machines (SVM) \cite{tenPas2018}, convolutional neural networks (CNN) \cite{7759114}, and other techniques to classify whether a grasp is antipodal or other engineered grasp representation features. Some methods directly process sensor outputs, such as depth images, and include representations of the grasp pose in their models \cite{DBLP:conf/rss/MahlerLNLDLOG17} \cite{Aktas2019DeepDG} \cite{9307474} \cite{46985}. Other approaches process the point cloud of the object, either directly \cite{cont_geo_aware_merwe}, as a voxel grid \cite{breyer2020volumetric}, or after a point cloud representation of the gripper has been integrated into the scene \cite{mousavian2019graspnet}.

Our dataset provides a benchmark for 6-DoF grasp success prediction with data collected on a parallel-jaw gripper. This section details our approach to training and evaluating a neural network for this task. We aim to demonstrate that achieving high prediction accuracy on our dataset is challenging, highlighting its value as a benchmark for future research. The code for reproducing the results is publicly available\footnote{\href{https://github.com/Jason-Toskov/Real-World-Object-and-Grasping-Dataset}{https://github.com/Jason-Toskov/Real-World-Object-and-Grasping-Dataset}}.

\begin{table*}[t]
    \centering
    \small
    \begin{NiceTabular}{|p{2.7cm}|*{10}{p{1cm}}|}\hline
        \Block{2-1}{\diagbox{\textbf{Success Status}}{\textbf{Objects}}} &
        \picdims[height=2.28cm]{1.03cm}{2.28cm}{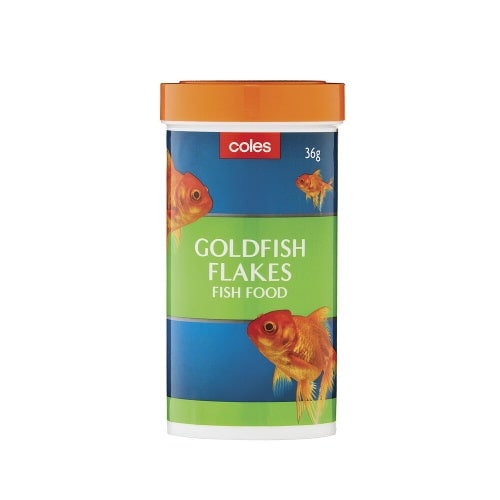} &
        \picdims[height=2.28cm]{1.03cm}{2.28cm}{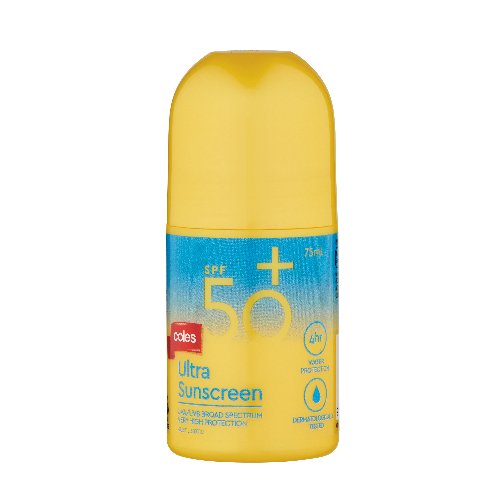} &
        \picdims[height=2.28cm]{1.03cm}{2.28cm}{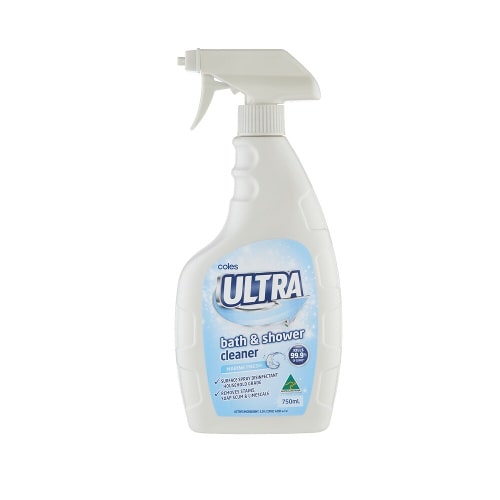} &
        \picdims[height=2.28cm]{1.03cm}{2.28cm}{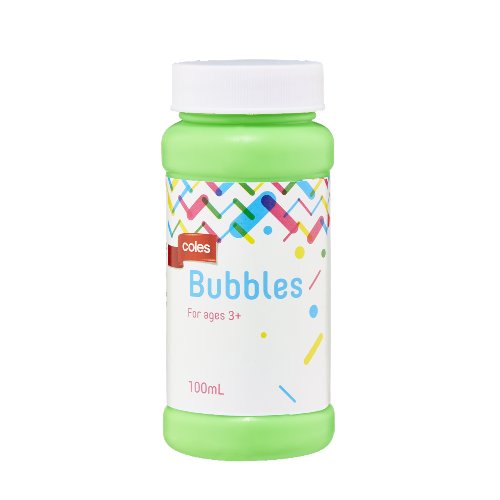} &
        \picdims[height=2.28cm]{1.03cm}{2.28cm}{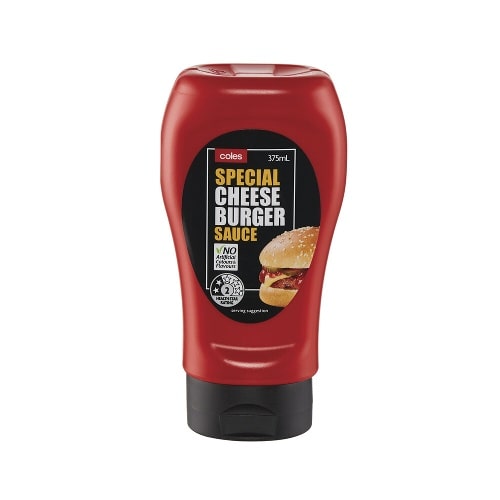} &
        \picdims[height=2.28cm]{1.03cm}{2.28cm}{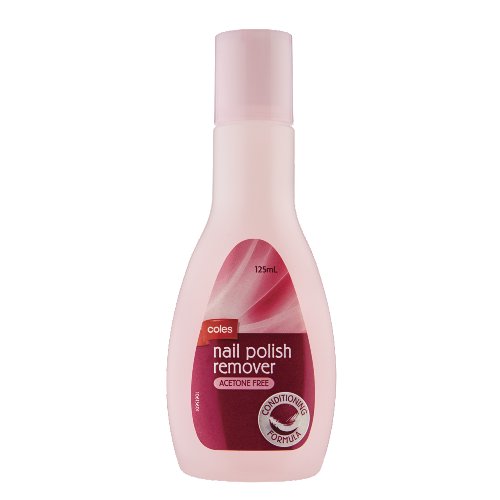} &
        \picdims[height=2.28cm]{1.03cm}{2.28cm}{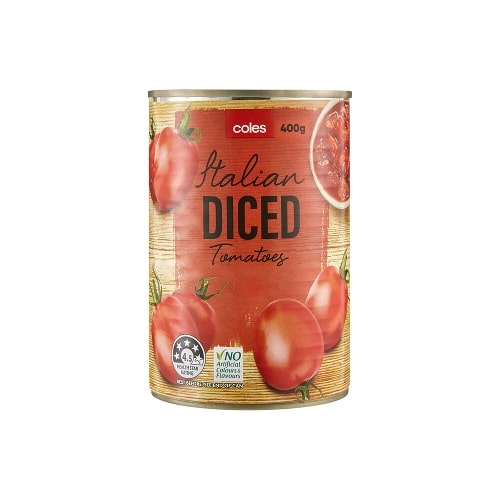} &
        \picdims[height=2.28cm]{1.03cm}{2.28cm}{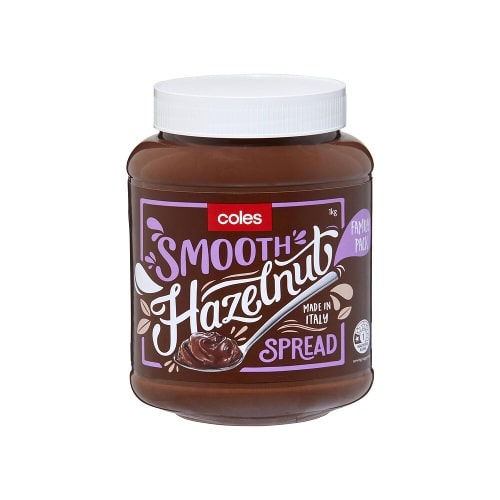} &
        \picdims[height=2.28cm]{1.03cm}{2.28cm}{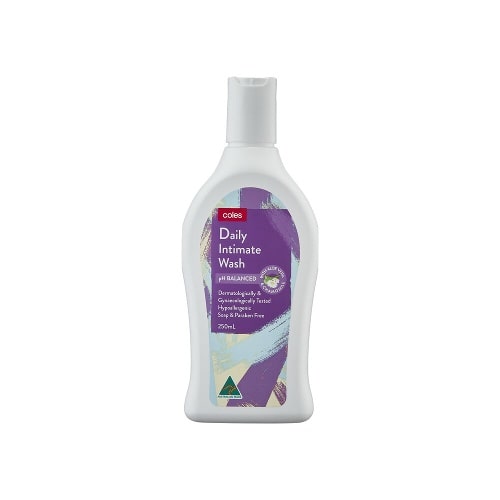} &
        \picdims[height=2.28cm]{1.03cm}{2.28cm}{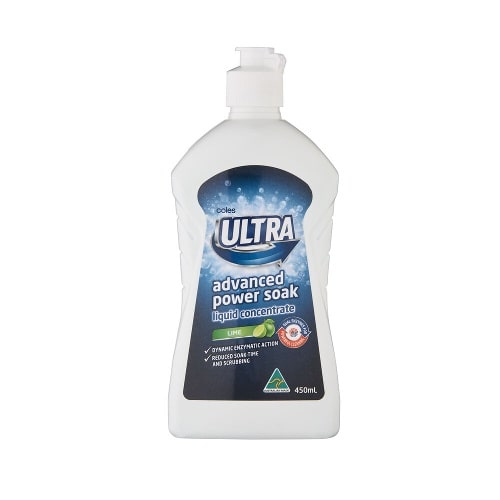} \\
        \cline{2-11}
        & \scriptsize Fish Flakes & \scriptsize Sunscreen Roll On & \scriptsize Bathroom Cleaner & \scriptsize Bubbles & \scriptsize Burger Sauce & \scriptsize NPR & \scriptsize Diced Tomatoes & \scriptsize Hazelnut Spread & \scriptsize Intimate Wash & \scriptsize Dishwash Liquid \\
        \hline
        Considered a success (model accuracy \%) &
        82.5 & 83.9 & 89.1 & 78.6 & 82.7 & 73.6 & 81.0 & 75.4 & 78.6 & 79.6 \\
        \hline
        Considered a failure (model accuracy \%) &
        74.5 & 69.8 & 84.2 & 74.4 & 79.6 & 75.8 & 73.0 & 71.4 & 74.6 & 66.6 \\
        \hline
    \end{NiceTabular}
\end{table*}
\vspace{-0.8cm}
\begin{table*}[ht!]
    \centering
    \small
    \begin{NiceTabular}{|p{2.7cm}|*{10}{p{1cm}}|}\hline
        \Block{2-1}{\diagbox{\textbf{Success Status}}{\textbf{Objects}}} &
        \picdims[height=2.28cm]{1.03cm}{2.28cm}{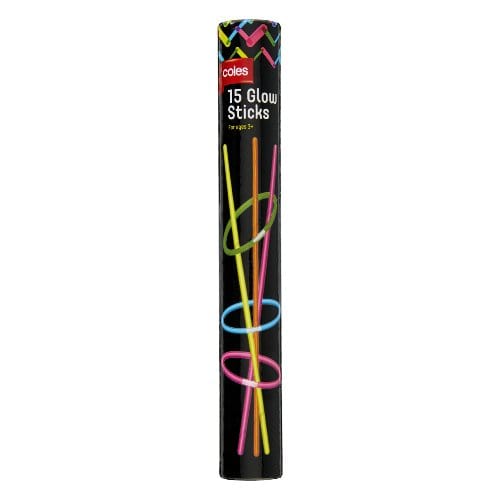} &
        \picdims[height=2.1cm]{1.03cm}{2.28cm}{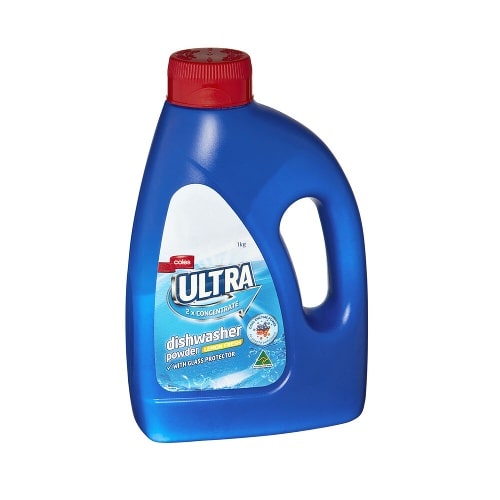} &
        \picdims[height=1.6cm]{1.03cm}{2.28cm}{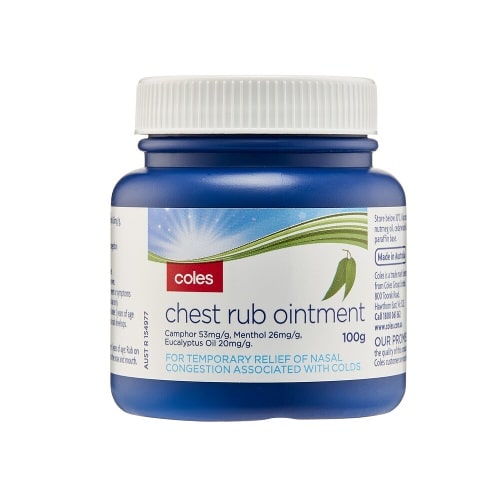} &
        \picdims[height=2.28cm]{1.03cm}{2.28cm}{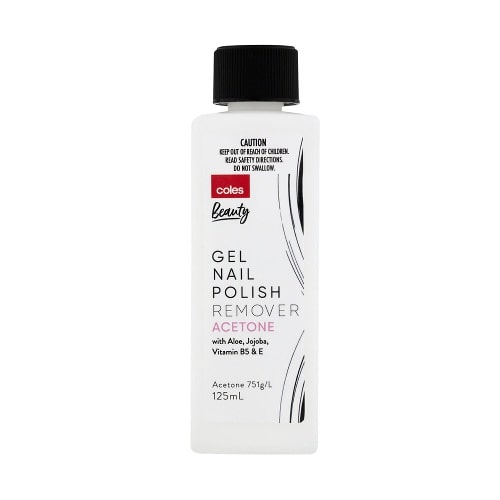} &
        \picdims[height=2.28cm]{1.03cm}{2.28cm}{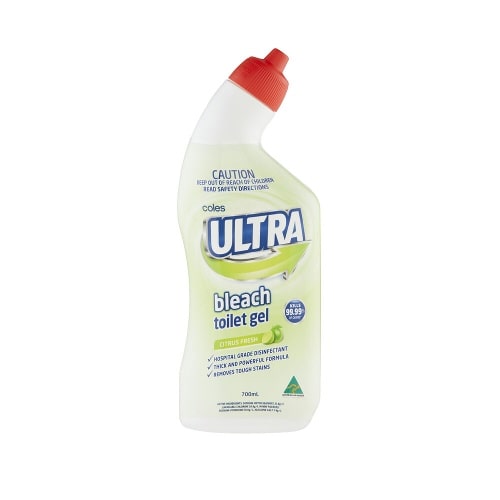} &
        \picdims[height=2.28cm, angle=90]{1.03cm}{2.28cm}{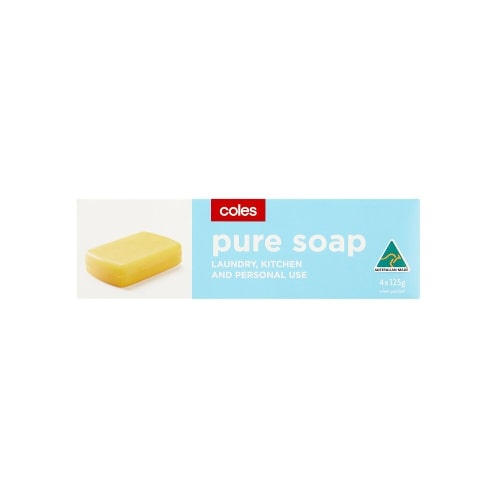} &
        \picdims[height=2.28cm]{1.03cm}{2.28cm}{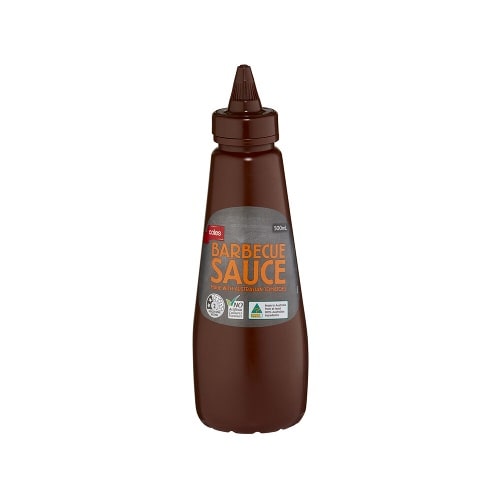} &
        \picdims[height=2.28cm, angle=90]{1.03cm}{2.28cm}{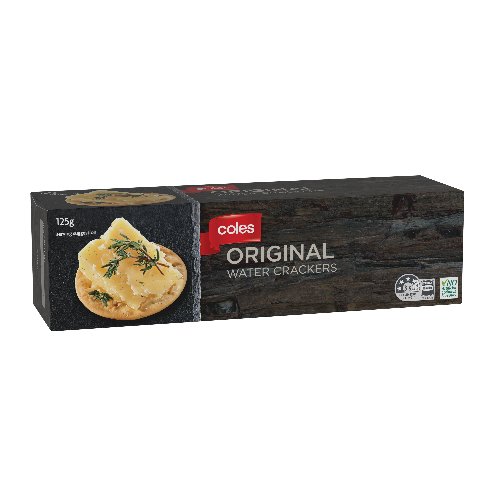} &
        \picdims[height=2.28cm]{1.03cm}{2.28cm}{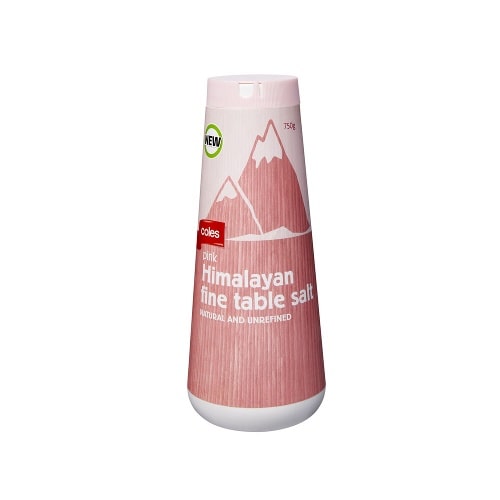} &
        \picdims[height=2.28cm]{1.03cm}{2.28cm}{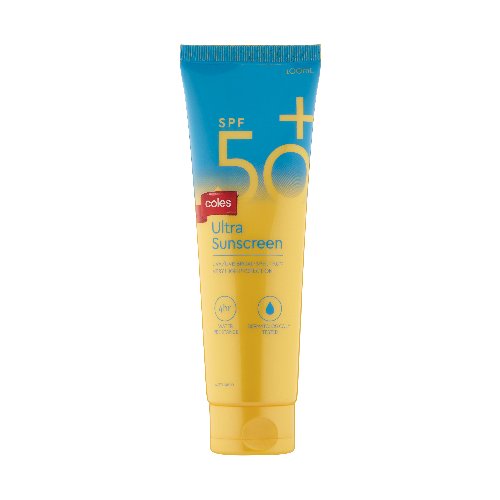} \\
        \cline{2-11}
        & \scriptsize Glowsticks & \scriptsize Dishwash Powder & \scriptsize Chest Ointment & \scriptsize Gel NPR & \scriptsize Toilet Cleaner & \scriptsize Soap Box & \scriptsize BBQ Sauce & \scriptsize Water Crackers & \scriptsize Salt & \scriptsize Sunscreen Tube \\
        \hline
        Considered a success (model accuracy \%) & 89.2 & 56.7 & 78.0 & 76.9 & 86.6 & 74.6 & 71.1 & 75.1 & 66.3 & 73.5 \\
        \hline
        Considered a failure (model accuracy \%) & 72.6 & 76.1 & 83.4 & 73.2 & 72.8 & 75.2 & 75.1 & 82.1 & 64.5 & 79.8 \\
        \hline
    \end{NiceTabular}
    \caption{5-fold cross-validation per-object classification accuracy for Grasp Success (Sec \ref{sec:grasp_success})}
    \label{tab:per_object_results}
    
\end{table*}

\subsection{Problem Definition}
Grasp success prediction is defined as a binary classification problem in which, given an input $x=\{P,G\}$ comprising of a point cloud of an object $P$ and a proposed grasp pose $G$, we aim to predict $y$, which is a binary label that classifies whether the grasp will be a success when executed. Each component is detailed below.

\textbf{Point Cloud}: The point cloud $P\in\mathbb{R}^{N\times6}$ consists of $N$ points, each with a corresponding location and normal vector oriented away from the center of the object. The number of points, $N$, varies due to differences in object sizes and the orientation and number of views used to construct the point cloud.  Texture information is disregarded for this experiment.

\textbf{Grasp}: The proposed grasp $G=\{p_{G}, \phi_{G}\}$ is defined by a grasp position $p_{G}\in\mathbb{R}^3$ and a quaternion $\phi_{G}\in\mathbb{H}$ of the gripper's center where it will close.

\textbf{Target}: The target $y=f_\theta(x)\in\{0,1\}$ defines whether the given grasp/cloud pair will result in a successful grasp. In our case, a successful grasp is indicated either by the success label or the stable success label as defined in section \ref{subsec:grasp_data_collection}.

The goal is to learn a function $f_\theta$ that maps the input $x$ to the target $y$ and minimises the disagreement between $y$ and the ground truth target $\hat{y}$. The optimization objective is:

\[ 
\theta_{opt}=\arg\min_{\theta\in\Theta}\mathbb{E}[\mathcal{L}(\hat{y},f_\theta(x))] 
\]

where $\mathcal{L}$ is the binary cross entropy loss and $\Theta$ are the model parameters.

\subsection{Point Cloud Pre-processing}
\label{subsec:grasp_pred_approach}

\begin{figure}[h!]
    \centering
    \includegraphics[width=0.45\textwidth]{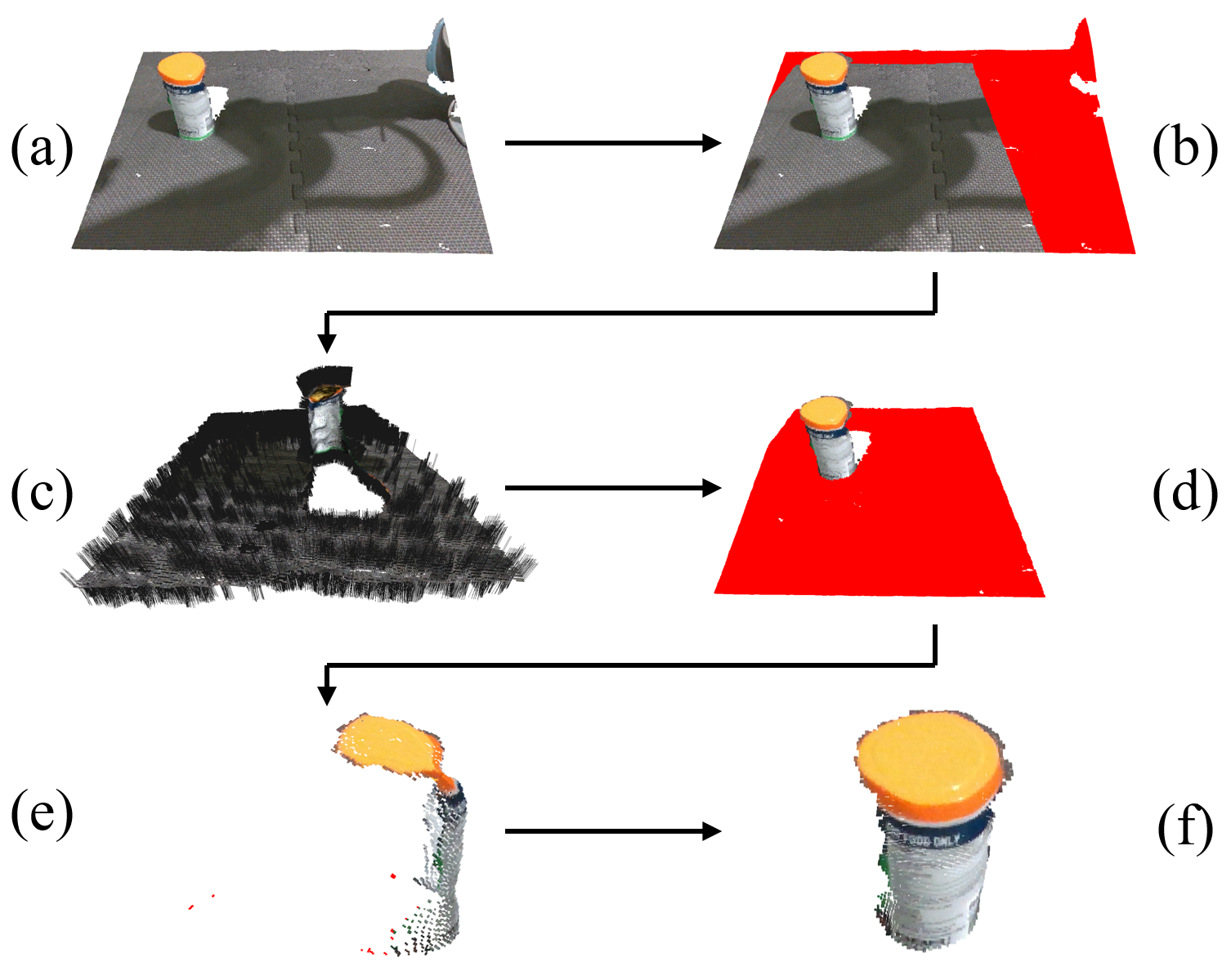}
    \caption{Point cloud pre-processing steps. Red color indicates points to be removed. (a) original point cloud, (b) workspace crop, (c) normal computation, (d) plane removal, (e) outlier removal, (f) downsampling.}
    \label{fig:grasp_preprocess}
    \vspace{-0.2cm}
\end{figure}

To prepare the point clouds, we apply several pre-processing steps to remove irrelevant points (e.g., ground plane or robot elements) and normalize the data. These steps, illustrated in Figure \ref{fig:grasp_preprocess}, include:

\begin{enumerate}[a)] 
\item \textbf{Original Point Cloud}
\item \textbf{Workspace Cropping}: Points outside the robot's workspace are removed to exclude irrelevant data.
\item \textbf{Normal Computation}: Normals for each point are computed and oriented away from the object's surface.
\item \textbf{Ground Plane Removal}: RANSAC is used to fit and remove the ground plane.
\item \textbf{Outlier Removal}: Statistical outlier filtering eliminates spurious points not captured by RANSAC.
\item \textbf{Downsampling}: Point clouds are randomly downsampled to 1024 points during training to maintain a consistent size.
\end{enumerate}

\subsection{Grasp Pose Representation}

We explore three different methods for representing the grasp pose:

\begin{enumerate}
    \item \textbf{Append Grasp Pose}: The grasp pose $G$, represented as a quaternion, is concatenated onto the first linear layer after the three point-set abstraction layers. The point cloud $P$ is translated so that the mean position of all points is at the origin.
    \item \textbf{Point Cloud in Gripper Frame}: The point cloud $P$ is transformed is transformed into the gripper's coordinate frame.This transformation removes the need to explicitly input the grasp pose $G$, as it can be inferred from the relative position of the object points.
    \item \textbf{Gripper as Point Cloud}: A 3D mesh model of the gripper is converted into a point cloud containing the same number of points (1024) as the object point cloud. This gripper point cloud is concatenated with the object point cloud, with a binary feature added to each point to distinguish whether it belongs to the gripper or the object. This method is similar to the approach used \cite{mousavian2019graspnet}. Note that the point cloud is also represented in the gripper frame similar to \textbf{Gripper as Point Cloud}.
\end{enumerate}

\begin{figure}[h!]
    \centering
    \begin{minipage}[t]{0.14\textwidth}
        \centering
        \fbox{\includegraphics[width=\textwidth]{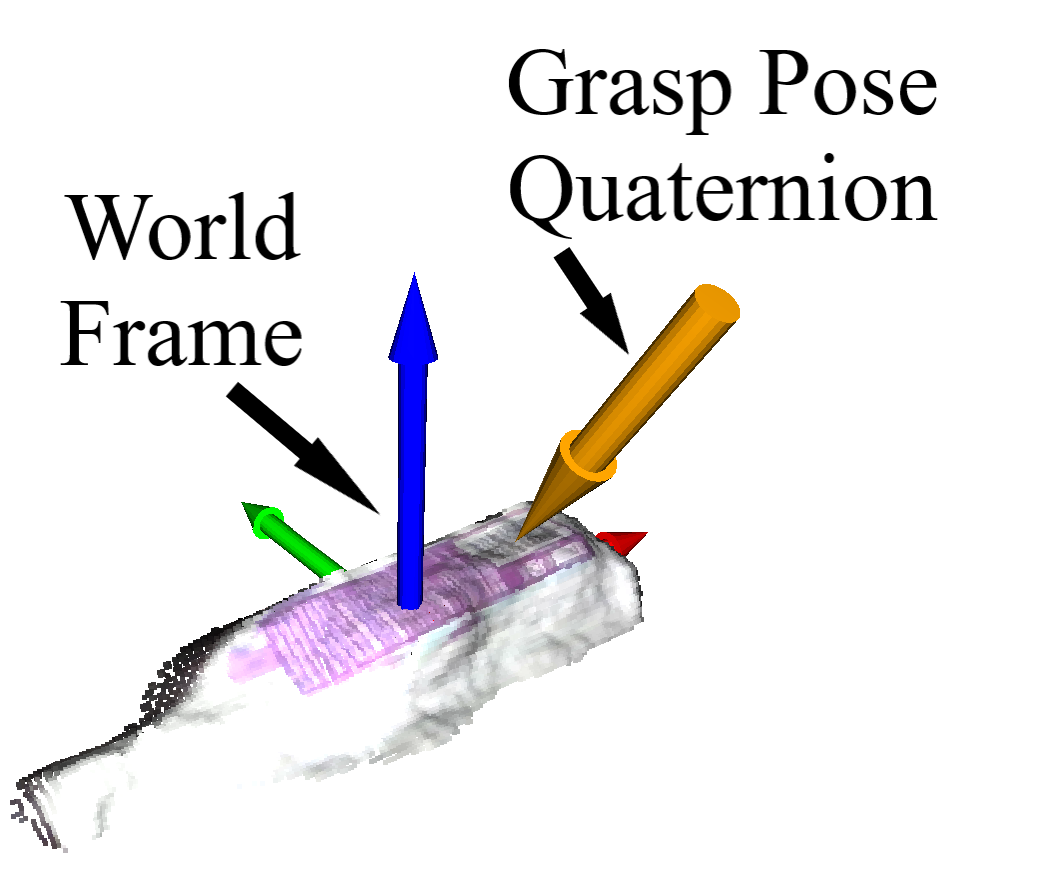}}
        \caption*{Append grasp pose} 
    \end{minipage}
    \hfill
    \begin{minipage}[t]{0.14\textwidth}
        \centering
        \fbox{\includegraphics[width=\textwidth]{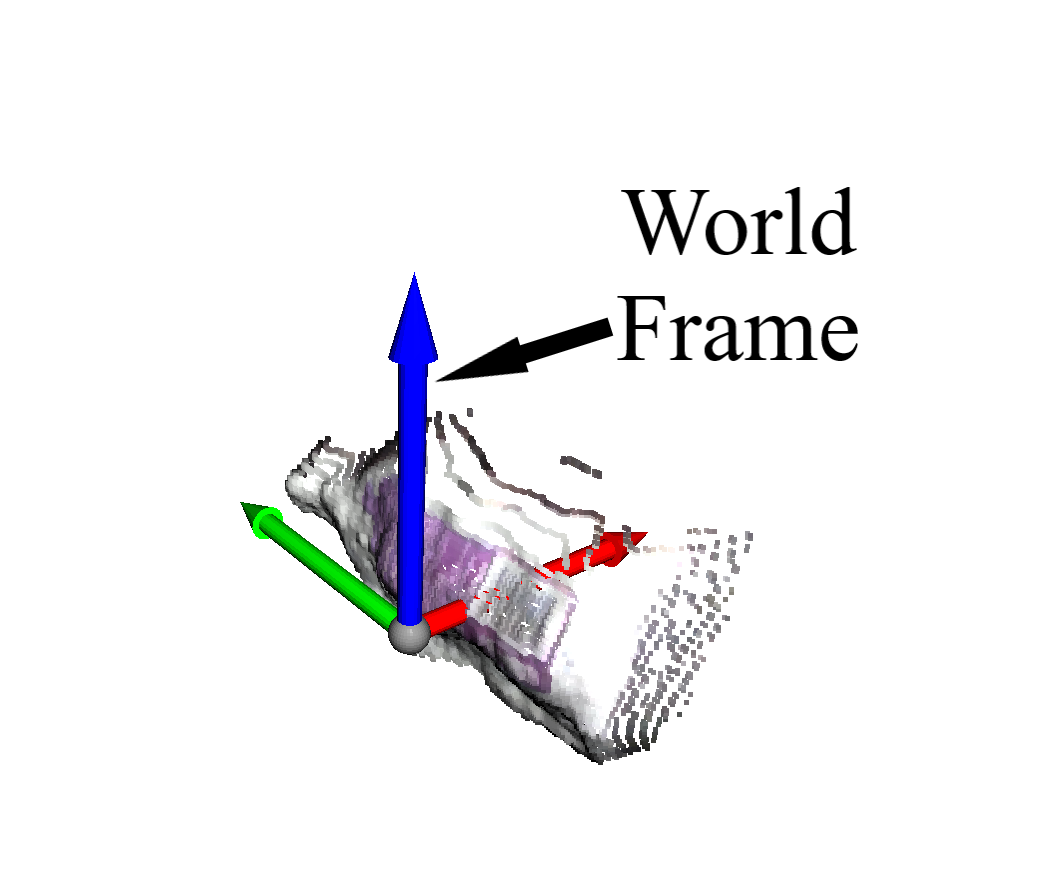}}
        \caption*{Point cloud in gripper frame}
    \end{minipage}
    \hfill
    \begin{minipage}[t]{0.14\textwidth}
        \centering
        \fbox{\includegraphics[width=\textwidth]{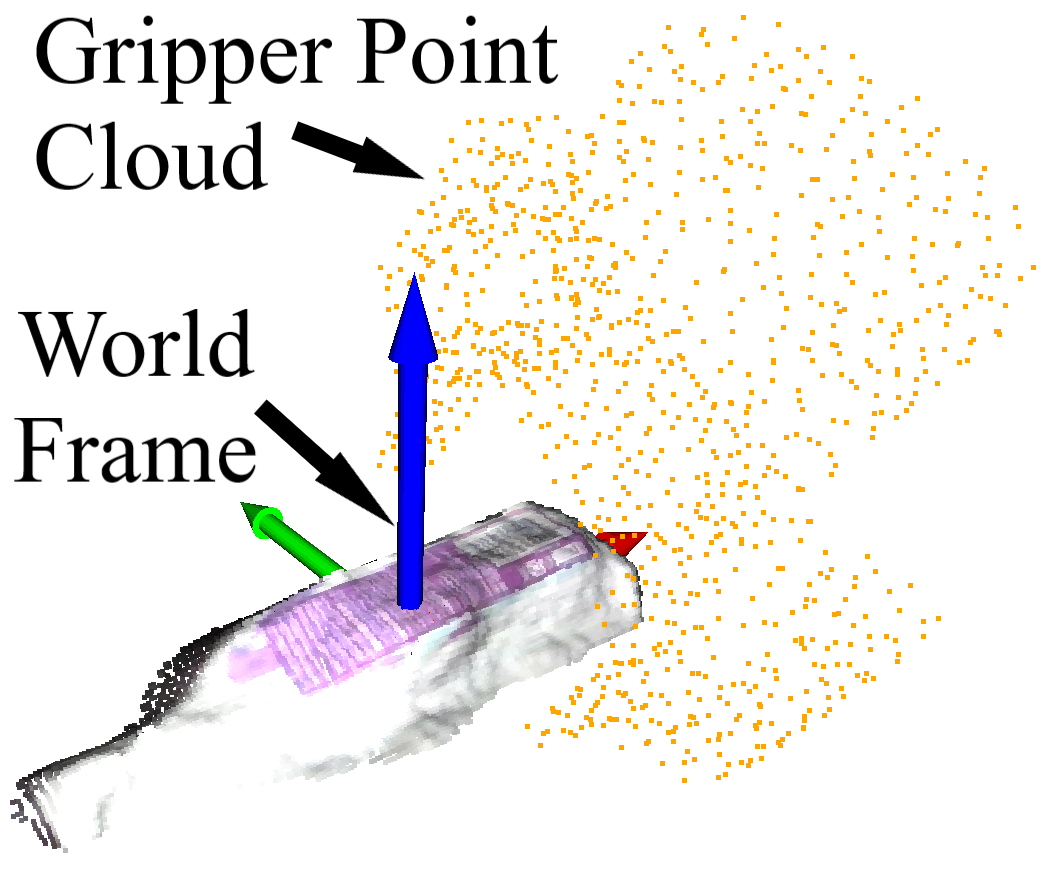}}
        \caption*{Gripper as point cloud} 
    \end{minipage}
    \caption{Grasp pose representations we explore for training}
    \label{fig:grasp_reps}
\end{figure}

These methods are illustrated in Figure \ref{fig:grasp_reps}. 

\subsection{Network Architecture and Training}

We use PointNet++ \cite{qi2017pointnetplusplus} as the base architecture for 6-DoF grasp success prediction due to its robustness in processing point clouds. Minor modifications were applied to adapt the model for our task. All models were trained for 200 epochs using the Adam optimizer with an initial learning rate of 0.001 (decayed by 0.7 every 20 epochs) and weight decay of 0.0001. The hyperparameters used to train the models were the same throughout all runs.

\section{Results}

\subsection{Grasp Success Prediction}
\label{sec:grasp_success}

We evaluated three different approaches for representing gripper poses in grasp success prediction using 5-fold cross-validation, focusing initially on binary grasp success (successful lift) while setting aside stability considerations. Table \ref{tab:unstable_is_success} presents the comparative performance of these approaches.

\begin{table}[h!]
    \centering
    \def\arraystretch{1.3}
    \normalsize
    \begin{tabular}{|l|c|} \hline
        \multicolumn{1}{|c|}{\textbf{Method}} &  \textbf{Accuracy ($\mu$ ± $\sigma$)} \\ \hline
        Append grasp pose & \%64.2 ± 1.7\\
        Cloud in gripper frame & \%73.4 ± 1.9 \\
        Gripper as point cloud & \%77.2 ± 2.8 \\ \hline
    \end{tabular}
    \caption{Grasp Success Prediction Accuracy}
    \label{tab:unstable_is_success}
\end{table}

The Gripper as Point Cloud method achieved the highest accuracy (77.2\%), demonstrating that explicit modeling of gripper geometry enhances the network's ability to understand grasp-object interactions. The Append Grasp Pose method performed least effectively (64.2\%), while the Point Cloud in Gripper Frame approach showed intermediate performance (73.4\%). 

Our findings indicate that simple quaternion representations lack sufficient context for establishing effective spatial relationships. The transformation of object point clouds into the gripper frame simplifies the learning task by creating a static grasp pose reference. Furthermore, representing the gripper as a point cloud facilitates learning of gripper-object interactions, leading to superior performance.

Analysis of per-object performance (Table \ref{tab:per_object_results}) reveals some variation across the 20 test objects. The Bathroom Cleaner demonstrated the highest prediction accuracy (89\% for successful grasps, 84\% for failures), while the Salt container proved most challenging (66\% for successful grasps, 64\% for failures). These variations underscore the dataset's complexity and its value as a benchmark for real-world grasp prediction algorithms.

\subsection{Stable Grasp Success Prediction}

We extended our analysis to consider ``stable success", grasps that maintain their hold through physical perturbations, a critical requirement for tasks such as object placement \cite{newbury2021learning} and in-hand manipulation \cite{toskov2023hand}. Among 1,500 grasps, 16\% succeeded in initial lifting but failed stability testing. Retraining the model to classify these cases as failures yielded updated accuracy scores (Table \ref{tab:stable_success_accuracy}).

\begin{table}[h!]
    \centering
    \def\arraystretch{1.3}
    \normalsize
    \begin{tabular}{|l|c|} \hline
        \multicolumn{1}{|c|}{\textbf{Method}} &  \textbf{Accuracy ($\mu$ ± $\sigma$)} \\ \hline
        Append grasp pose & \%67.3 ± 1.1\\
        Cloud in gripper frame & \%73.0 ± 1.3 \\
        Gripper as point cloud & \%75.2 ± 1.0 \\ \hline
    \end{tabular}
    \caption{Stable Grasp Success Prediction Accuracy}
    \label{tab:stable_success_accuracy}
\end{table}

Results were similar to the previous analysis, with \textbf{Gripper as Point Cloud} method maintaining superior performance (75.2\%) despite a 2 percentage point decrease in accuracy compared to prediction results for standard success prediction.

Notably, among the three alternatives for representing the grasp pose, representing the gripper as a point cloud still achieved the highest prediction accuracy, despite experiencing a 2 percentage point decrease. However, this decrease is relatively minor when compared to the percentage of unstable grasps in the dataset. The class imbalance in this problem, where stable successes are a minority, may have contributed to the slight reduction in accuracy.

An examination on the successful grasps that were not stable revealed that two factors may be in play. Large/heavy objects present challenges in identifying slip-resistant grasp poses. Moreover, we observed that rigid objects tend to pivot during gripper rotation, possibly due to material stiffness impeding secure grasps.


\subsection{Comparing Success vs Stable Success Prediction}

\begin{figure}[h!]
    \centering
    \includegraphics[width=0.5\textwidth]{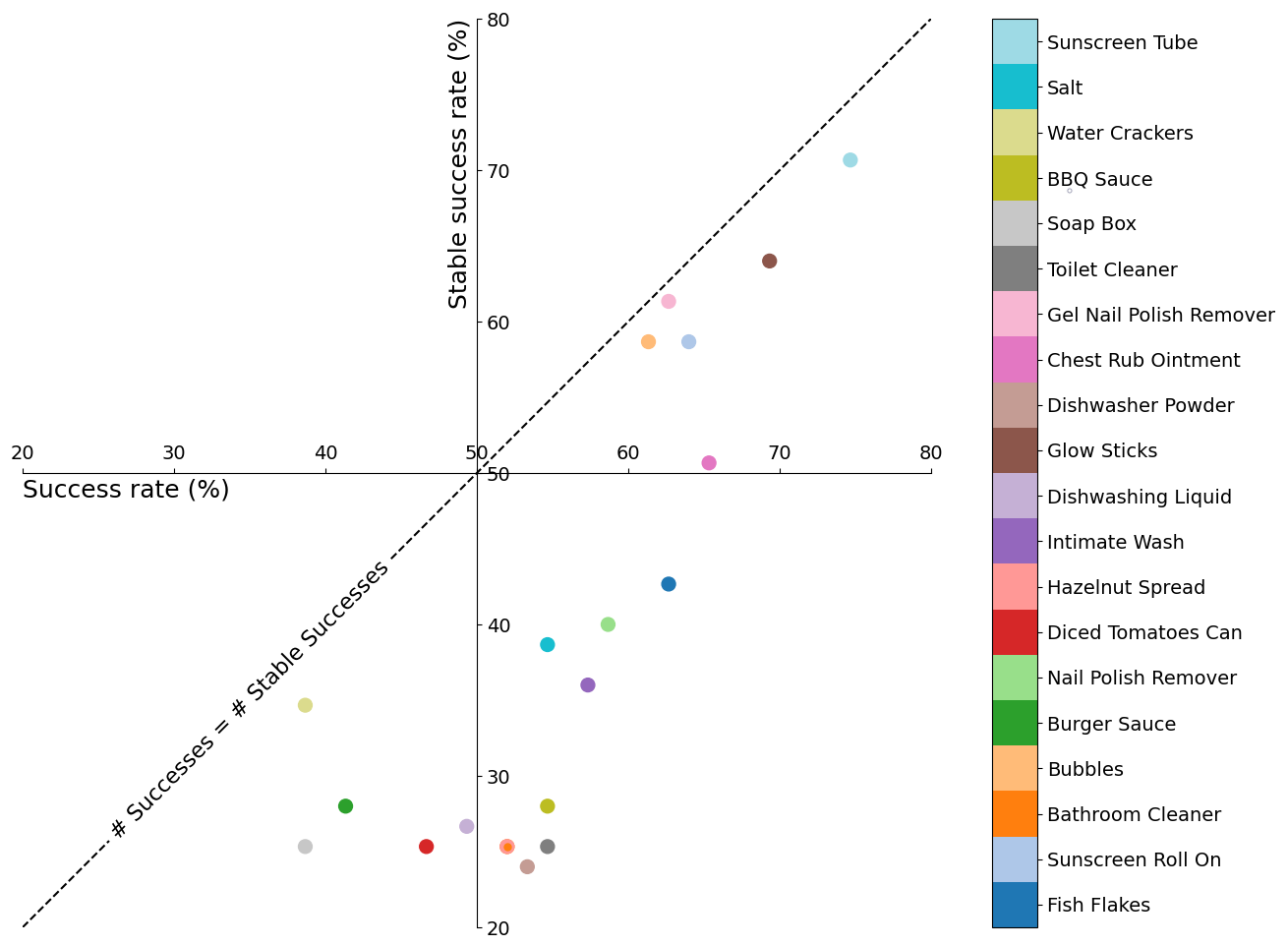}
    \caption{Success rate vs stable success rate of grasped objects.}
    \label{fig:success_vs_robust}
\end{figure}

Figure \ref{fig:success_vs_robust} compares success rates with stable success rates across test objects using the best-performing grasp pose representation, \textbf{Gripper as Point Cloud}. 

The data reveals that predicting stable grasp success is a more difficult task than predicting standard grasp success across all objects. Notably, the top four objects with the most significant performance drop in stable grasp success prediction all weighed over 600g, highlighting object weight as a crucial factor in grasp stability. Future research can potentially incorporate object weight as an additional indicator to improve prediction performance.

These findings demonstrate that models can effectively learn to identify unstable grasps without substantial performance degradation. We advocate for incorporating stability considerations in future grasp success prediction algorithms to enhance their practical utility in real-world applications.

\section{Conclusion}
\label{sec:conclusion}

This paper introduces the Supermarket-6DoF dataset, comprising 1,500 grasp attempts executed on a real robot. While existing datasets predominantly rely on simulated trials or analytical quality metrics, our work provides authentic grasp data executed on physical hardware. Although our dataset is smaller than comparable real-world grasping datasets (Pinto et al. \cite{Pinto2015SupersizingSL} and Levine et al. \cite{doi:10.1177/0278364917710318}), it offers unique advantages through 6-DoF grasp poses and stability labels. Each grasp attempt includes single-view RGB and depth images alongside corresponding point clouds, providing rich sensory information for learning algorithms. Our focus on common supermarket objects, complete with available 3D models, ensures the dataset's practical relevance while maintaining accessibility for researchers. The public availability of our dataset facilitates reproducible research.

We also present an analysis of grasp pose representations for predicting grasp success using our dataset. Our results show that explicitly modeling the gripper as a point cloud significantly outperforms the conventional approach of appending grasp poses to fully connected layers. The prediction performance variations across different objects and grasp stability highlight the complexity of real-world grasping.

Looking ahead, we believe that real-world datasets like Supermarket-6DoF will be instrumental in developing robust grasping algorithms capable of handling the full complexity of practical manipulation tasks.

\balance
\bibliographystyle{unsrt} 
\bibliography{ref.bib}

\end{document}